 \documentclass[authoryear,preprint,review,11pt]{elsarticle}

\usepackage{amssymb}
\usepackage{amsmath}
\usepackage{comment}
\usepackage{algorithm}
\usepackage{algorithmic}
\usepackage{amsmath}
\usepackage{amssymb}
\usepackage{multirow}
\usepackage{multicol}
\usepackage{amsmath,amssymb,amsfonts}
\usepackage{xcolor}

\journal{Pattern Recognition}
\setcitestyle{numbers}  % added
\setcitestyle{square}

\begin{document}

\begin{frontmatter}



% \title{Unbiased Model Prediction Without Protected Attribute Information}
\title{Unbiased Model Prediction Without Using Protected Attribute Information}


\author[iiitaddress,iitjaddress]{Puspita~Majumdar\fnref{iiitfootnote}}
\author[iitjaddress]{Surbhi Mittal\fnref{iitfootnote}}
\author[iiitaddress,iitjaddress]{Saheb Chhabra\fnref{iiitfootnote}}
\author[iitjaddress]{Mayank Vatsa\fnref{iitfootnote}}
\author[iitjaddress]{Richa Singh\fnref{iitfootnote}\corref{mycorrespondingauthor}}
\address[iiitaddress]{IIIT-Delhi, New Delhi, India, 110020}
\address[iitjaddress]{IIT Jodhpur, India, 342037}
\fntext[iiitfootnote]{email: \{pushpitam, sahebc\}@iiitd.ac.in}
\fntext[iitfootnote]{email: \{mittal.5, mvatsa, richa\}@iitj.ac.in}
\cortext[mycorrespondingauthor]{Corresponding author}

\begin{comment}

\author{} %% Author name

\affiliation{organization={},%Department and Organization
            addressline={}, 
            city={},
            postcode={}, 
            state={},
            country={}}

\end{comment}

%% Abstract
\begin{abstract}
The problem of bias persists in the deep learning community as models continue to provide disparate performance across different demographic subgroups. Therefore, several algorithms have been proposed to improve the fairness of deep models. However, a majority of these algorithms utilize the protected attribute information for bias mitigation, which severely limits their application in real-world scenarios. To address this concern, we have proposed a novel algorithm, termed as \textbf{Non-Protected Attribute-based Debiasing (NPAD)} algorithm for bias mitigation, that does not require the protected attribute information. The proposed NPAD algorithm utilizes the auxiliary information provided by the non-protected attributes to optimize the model for bias mitigation. Further, two different loss functions, \textbf{Debiasing via Attribute Cluster Loss (DACL)} and \textbf{Filter Redundancy Loss (FRL)} have been proposed to optimize the model for fairness goals. Multiple experiments are performed on the LFWA and CelebA datasets for facial attribute prediction, and a significant reduction in bias across different gender and age subgroups is observed.
\end{abstract}

\begin{comment}
    \begin{graphicalabstract}
\end{graphicalabstract}
\end{comment}

\begin{highlights}
\item Propose a novel Non-Protected Attribute-based Debiasing (NPAD) algorithm for unbiased model prediction. The proposed NPAD algorithm does not require apriori knowledge of the protected attributes for optimizing the model for fairness goals.
\item Propose two novel loss functions, Debiasing via Attribute Cluster Loss (DACL) and Filter Redundancy Loss (FRL) for fair model training. DACL provides supervision to the model to learn class-specific information for bias mitigation and FRL helps boost the performance by enabling the model to learn non-redundant features.
\item Introduce a novel metric, Overall Performance Equality (OPE) to fairly measure the model performance across different subgroups.
\item Extensive experiments are performed on the LFWA and CelebA datasets to demonstrate the effectiveness of the proposed NPAD algorithm for bias mitigation in facial attribute prediction and comparisons are performed with existing algorithms.
\end{highlights}

%% Keywords
\begin{keyword}
Bias, Fairness, Unbiased Predictions, Unknown Protected Attribute, Facial Attribute Prediction, Deep Learning

\end{keyword}

\end{frontmatter}



\section{Introduction}
\label{sec:intro}

The prevalence of bias in deep learning community has gained significant attention in recent years. Several instances of \textit{bias} in model predictions have been observed \cite{buolamwini2018gender}, \cite{majumdar2021unravelling}, \cite{karkkainen2021fairface}. This biased behavior of deep models leads to a disparate impact by negatively affecting certain gender, race, or age subgroups. For instance, Twitter's image-cropping algorithm has been recently found to favor young and lighter-skinned people over others \cite{Twitter}. The presence of such bias in our day-to-day applications leads to unfair outcomes for certain subgroups, thereby reinforcing existing societal biases.

\begin{figure*}[!t]
\centering
\includegraphics[scale = 0.35]{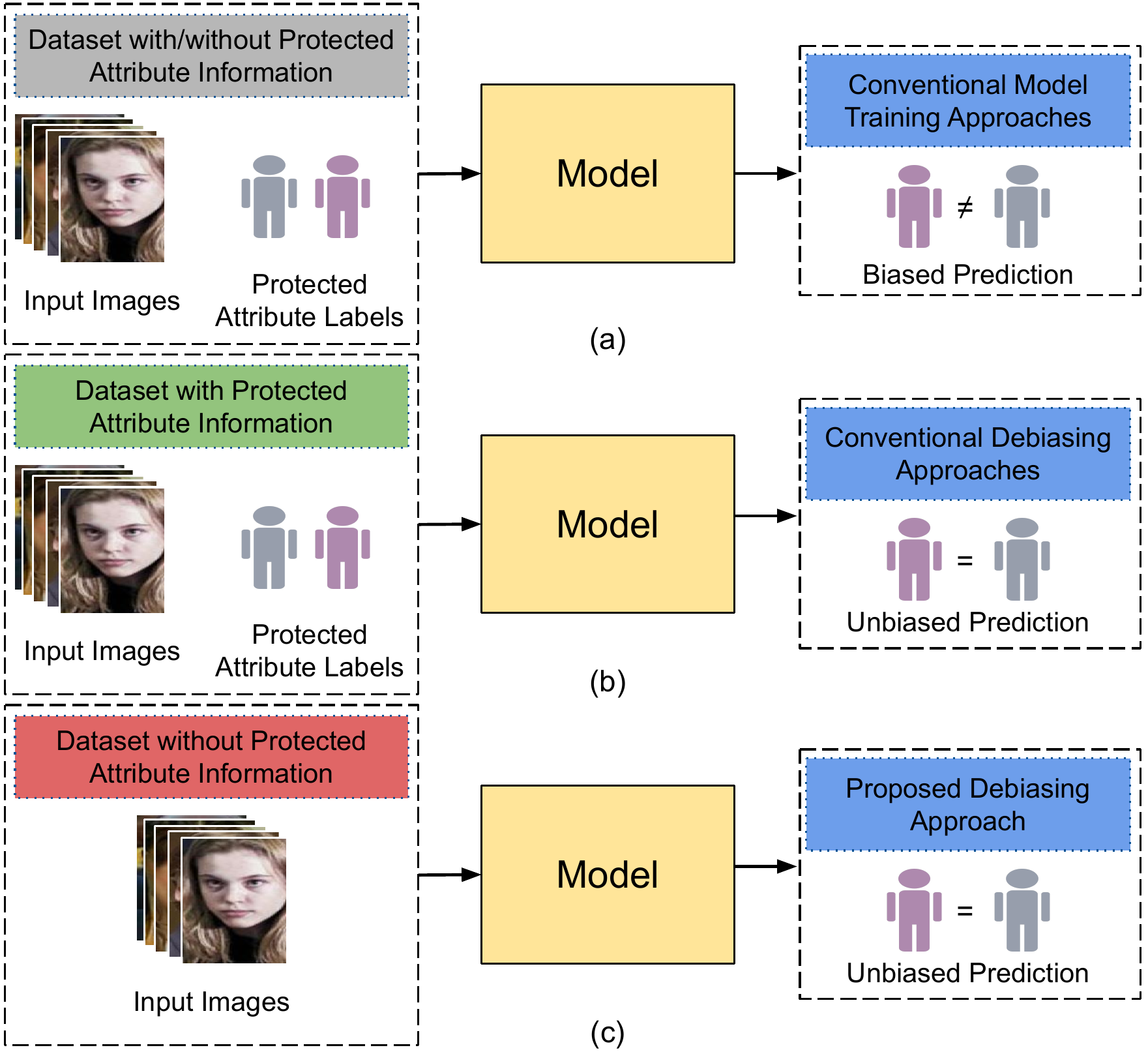}
\caption{Illustration of model training using (a) conventional approaches, (b) conventional debiasing approaches, and (c) proposed Non-Protected Attribute-based Debiasing (NPAD) algorithm. The proposed algorithm does not require the protected attribute information for bias mitigation.}
\label{fig:VisualAbstract}
\end{figure*}

Research has shown that the biased behavior of deep models is due to unknowingly learning and encoding the features related to the protected attributes \cite{pezeshki2020gradient}, \cite{serna2021insidebias}, \cite{9650887}. For example, a smiling/not-smiling prediction model may encode features related to gender, leading to biased outcomes (Figure \ref{fig:VisualAbstract}(a)). A large number of approaches have been proposed for mitigating bias in the deep learning \cite{majumdar2020subgroup}, \cite{gong2021mitigating}, \cite{kolling2022mitigating}, \cite{agarwal2022does}, \cite{cao2022fair}. However, most approaches focus on using the information of the protected attributes (Figure \ref{fig:VisualAbstract}(b)). The availability of such information is not always feasible in the real world. In some countries, it is against the law to disclose sensitive information such as gender or race to circumvent discrimination \cite{SHRM}. 

%\begin{figure}[!t]
%\centering
%\includegraphics[width=2.5in]{myfigure}
% where an .eps filename suffix will be assumed under latex, 
% and a .pdf suffix will be assumed for pdflatex; or what has been declared
% via \DeclareGraphicsExtensions.
%\caption{Simulation results for the network.}
%\label{fig_sim}
%\end{figure}

To address the concern of bias mitigation when protected attribute information is unavailable, in this research, we propose a novel algorithm, termed as \textbf{Non-Protected Attribute-based Debiasing (NPAD)} algorithm. The proposed NPAD algorithm does not require apriori knowledge of the protected attributes for optimizing the model for fairness goals (Figure \ref{fig:VisualAbstract}(c)). Instead, it utilizes the auxiliary information provided by the non-protected attributes for bias mitigation. The objective of NPAD is two-fold \textit{(i) intelligently select the non-protected attributes} and \textit{(ii) use the selected attributes to optimize the model for bias mitigation}. Non-protected attribute selection is done based on the disparity of the attribute prediction model across other non-protected attributes. After non-protected attribute selection, the model is optimized using the proposed \textbf{Debiasing via Attribute Cluster Loss (DACL)} and \textbf{Filter Redundancy Loss (FRL)}. DACL provides supervision to the model to learn class-specific information for bias mitigation and FRL helps boost the performance by enabling the model to learn non-redundant features. Additionally, we propose a new fairness metric, \textbf{Overall Performance Equality (OPE)} to fairly measure the model performance across different subgroups. The effectiveness of NPAD is demonstrated for bias mitigation in facial attribute prediction through multiple case studies and comparison with existing algorithms~\cite{nam2020learning}, \cite{lahoti2020fairness}.

\begin{figure*}[t]
\centering
\includegraphics[scale=0.24]{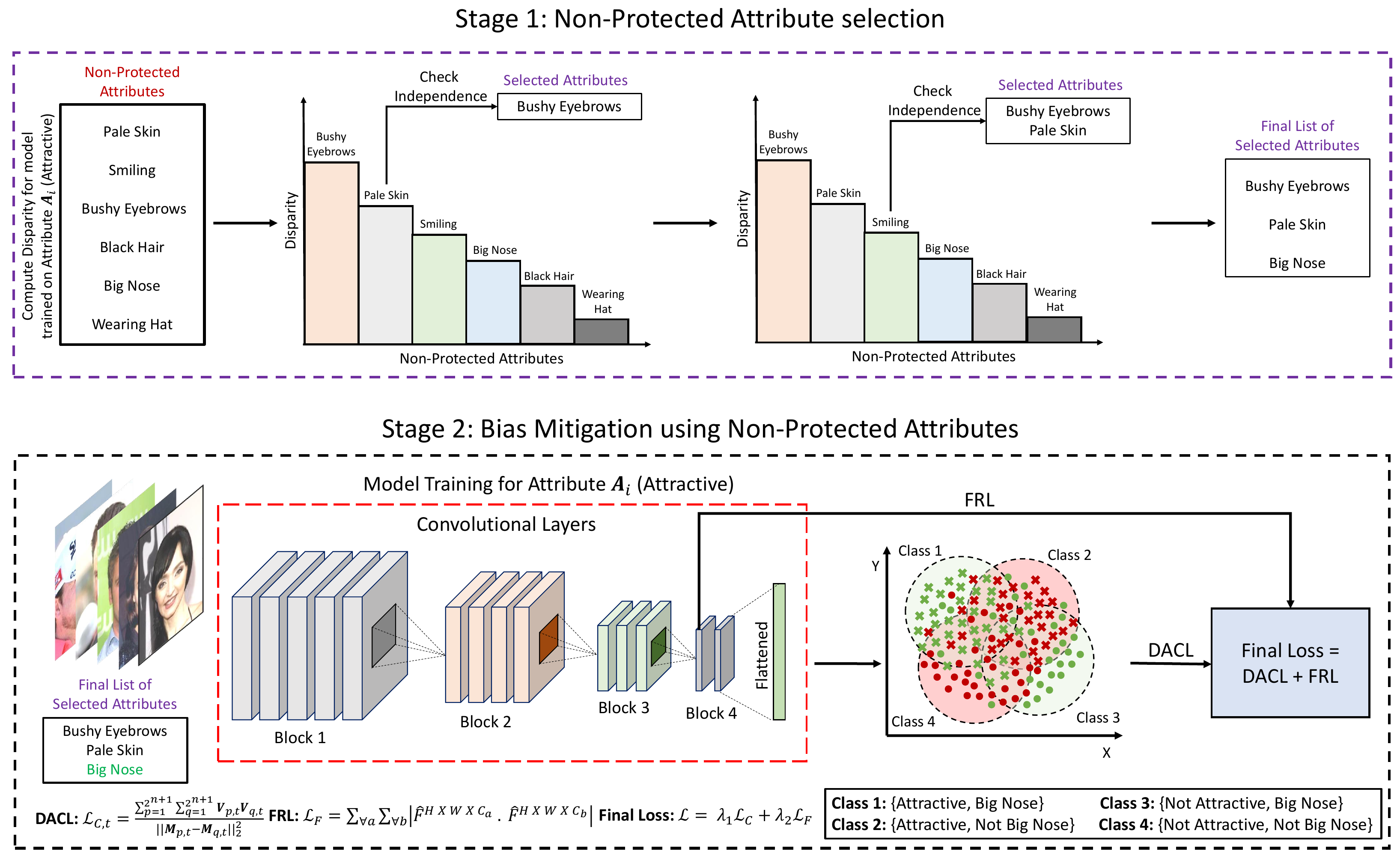}
\caption{Block diagram illustrating model training using the proposed NPAD algorithm. In the first stage, disparity is computed by evaluating the model trained for predicting attribute $\mathbf{A}_i$ across the classes of other non-protected attributes. Next, the non-protected attributes are selected based on the disparity and independence. In the second stage, model is optimized using the proposed DACL and FRL functions for bias mitigation.}
\label{fig:BlockDiagram}
\end{figure*}

% They employed a joint dynamic weight adjustment for the same.

% \cite{kim2019learning} proposed a regularization loss to unlearn the bias through minimization of mutual information present in the feature embeddings. A variety of algorithms have also been proposed to mitigate bias in pre-trained and black-box models. \cite{nagpal2020diversity} used \textit{diversity blocks} on top of existing models to debias their output.

% \cite{kim2019multiaccuracy} proposed a post-processing technique termed as MULTIACCURACY BOOST, which improves the performance across identifiable subgroups. 

% \cite{ramaswamy2021fair} proposed a technique for debiasing the data in the latent space by generating images using GANs and balancing data across all the attributes in the dataset.

\section{Related Work}

A wide range of algorithms have been proposed for mitigating bias in attribute prediction, including generative, adversarial, and deep learning-based approaches. Generative approaches include the oversampling of minority classes for managing the class imbalance problem. In this direction, Mullick et al. \cite{mullick2019generative} and Ramaswamy et al. \cite {ramaswamy2021fair} use convex generators near the peripheries of the classes and generative adversarial networks (GANs) for generating images of the minority class, respectively. Adversarial approaches for bias mitigation include learning a uniform perturbation for the input dataset to generate a transformed dataset \cite{majumdar2020subgroup}. The transformed dataset produces unbiased outcomes when given as input to the model. In deep learning-based approaches, a Multi-Task CNN is proposed by  Das et al. \cite{das2018mitigating}, which mitigates bias by joint classification of race, age, and gender attributes. Huang et al. \cite{huang2025debiasing} mitigate bias from redundant feature correlations caused by selection bias, specifically for multi-view clustering. Liu et al. \cite{liu2024unbiased} present an unbiased self-supervised representation learning method that utilizes an unbiased graph structure to reduce the impact of noise on model prediction.

Chuang et al. \cite{chuang2021fair} proposed a \textit{mixup} data augmentation technique for debiasing which regularizes the model based on interpolated samples between different demographic groups. Park et al. \cite{park2021learning} used a feature disentanglement approach in which they separated the data representation into subspaces based on the target, protected, and other attributes. By utilizing the protected attribute information, the authors reported improved fairness. Kolling et al. \cite{kolling2022mitigating} generated new annotations from existing ones and utilized them for training multiple models on the different annotations. They showcased that label diversity improves the overall fairness of the models. Recently, a work \cite{qraitem2023bias} introduced a new class-conditioned sampling method, termed as \textit{Bias Mimicking} for fair model training. The proposed sampling method ensures that the model is exposed to the entire distribution per epoch without repeating samples. In \cite{zeng2023fairness}, the authors used adversarial robustness of individual training samples to re-weigh them during training for fairness-aware training process. Majumdar et al. \cite{majumdar2023uniform} propose the Uniform Misclassification Loss to train models for unbiased outcomes by penalizing the worst-performing subgroup. Wang et al. \cite{wang2025marginal} propose a marginal debiased network that utilizes a margin penalty for mitigating bias caused by spurious correlations.

% % Added Surbhi
Some research work focusing on mitigating bias without protected attribute information include Adversarially Reweighted Learning (ARL)~\cite{lahoti2020fairness}. In ARL, the authors propose to focus on computationally-identifiable regions of errors to improve model performance for protected groups. Another work in this direction is Learning from failure (LfF)~\cite{nam2020learning}. The LfF algorithm trains two neural networks and amplifies biases in the first model. The second model is trained such that it rejects the predictions made by the first (biased) network.  

The majority of existing algorithms for bias mitigation in attribute prediction utilize the information of protected attributes. However, the availability of protected attributes is either not feasible or permitted in many real-world applications due to privacy or legal issues. In this research, the proposed NPAD algorithm does not require protected attribute information for bias mitigation.

% - Debiasing weighted multi-view k-means clustering based on causal regularization \cite{huang2025debiasing}
% - Marginal debiased network for fair visual recognition \cite{wang2025marginal}
% - Enhancing performance of vision transformers on small datasets through local inductive bias incorporation \cite{akkaya2024enhancing}
% - Unbiased and augmentation-free self-supervised graph representation learning \cite{liu2024unbiased}
% - Uniform misclassification loss for unbiased model prediction \cite{majumdar2023uniform}

\section{Non-Protected Attribute-based Debiasing}
Existing studies have shown that mitigating bias with respect to protected attributes does not always ensure fair prediction, as the correlation of non-protected attributes with protected attributes eventually leads to biased predictions \cite{wang2019balanced}. For example, geographic data (e.g., zip code) is highly correlated with race. Thus, a model indirectly encodes the features related to the protected attribute (race) through the non-protected attribute (zip code), resulting in biased outcomes \cite{Datta}. This research positively exploits this correlation and uses it for bias mitigation when the protected attribute information is unavailable.   

\subsection{Problem Formulation}  
Given a dataset $\mathbf{X}$ and the corresponding non-protected attribute set $\mathbf{A} = \{\mathbf{A}_1, \mathbf{A}_2, ..., \mathbf{A}_k\}$, where, $\mathbf{A}_i$ is a binary attribute in set $\mathbf{A}$, i.e., $\mathbf{A}_i \in [0,1]$. The aim is to increase the fairness of the model for predicting attribute $\mathbf{A}_i$ across the unknown protected attributes (e.g., \textit{gender}, \textit{age}) by intelligently selecting the non-protected attributes from set $\mathbf{A}$ and utilizing them to optimize the model. 

To address the problem, we have proposed a novel \textbf{Non-Protected Attribute-based Debiasing (NPAD)} algorithm. The proposed NPAD algorithm comprises of two stages \textit{(i) non-protected attribute selection} and \textit{(ii) bias mitigation using the selected non-protected attributes}. Figure \ref{fig:BlockDiagram} shows the block diagram of bias mitigation using the proposed NPAD algorithm. The details of the NPAD algorithm are discussed in the following subsections.

\subsection{Non-protected Attribute Selection}
Consider $\phi_i$ as a trained model for predicting attribute $\mathbf{A}_i$. One option to select the non-protected attributes is to randomly select the attributes from set $\mathbf{A}$ (excluding attribute $\mathbf{A}_i$). However, since we are utilizing the biased behavior of model $\phi_i$ across non-protected attributes to mitigate bias across the protected attributes, such random selection is not feasible. If the model $\phi_i$ is fair across the classes of the selected attributes, optimizing the model $\phi_i$ for fairness goals will be futile. Therefore, it is important to select the attributes for which the model $\phi_i$ shows high disparity. For this purpose, the trained model $\phi_i$ is evaluated across the classes of other non-protected attributes, and the disparity in model prediction is computed corresponding to each attribute. Let $\mathbf{D}_i = \{D_{i,1}, D_{i,2}, ..., D_{i,k-1}\}$ denotes the disparity set, where $D_{i,j}$ represents the disparity in model prediction across the classes of the non-protected attribute $\mathbf{A}_j$. The disparity $D_{i,j}$ is computed by taking the standard deviation of the classification accuracy across the classes of $\mathbf{A}_j$. The non-protected attribute for which the model $\phi_i$ shows the highest disparity is selected as the first attribute.

% For selecting the first non-protected attribute for which the model $\phi_i$ shows the highest disparity, $\mathbf{D}_i$ is sorted in descending order, and the corresponding attribute is selected.

We sort the disparity set $D_{i,j}$ in the descending order to pick the non-protected attribute for which the model $i$ shows the largest disparity. Let $\mathbf{S}_i$ is the sorted disparity set (descending). Ideally, if the selected non-protected attribute is fully correlated with the protected attributes, optimizing the model for the selected attribute will effectively mitigate the bias in model prediction and improve the fairness across the protected attributes. However, in our case, we do not have the protected attribute information. Therefore, while selecting other non-protected attributes, it is important that the selected attributes are independent of each other as it will reduce the redundancy in the selected attributes and increase the probability of obtaining non-protected attributes correlated with the protected attributes. Thus, the second non-protected attribute selected based on $\mathbf{S}_i$ must be independent of the first attribute. In this research, the $\chi^2$ test of independence is used to check the independence of the selected attributes. The above process of selecting the non-protected attributes is repeated for all the attributes in set $\mathbf{A}$ (excluding attribute $\mathbf{A}_i$) based on $\mathbf{S}_i$ and $p$-value (for independence).

\begin{figure*}[]
\centering
\includegraphics[scale = 0.4]{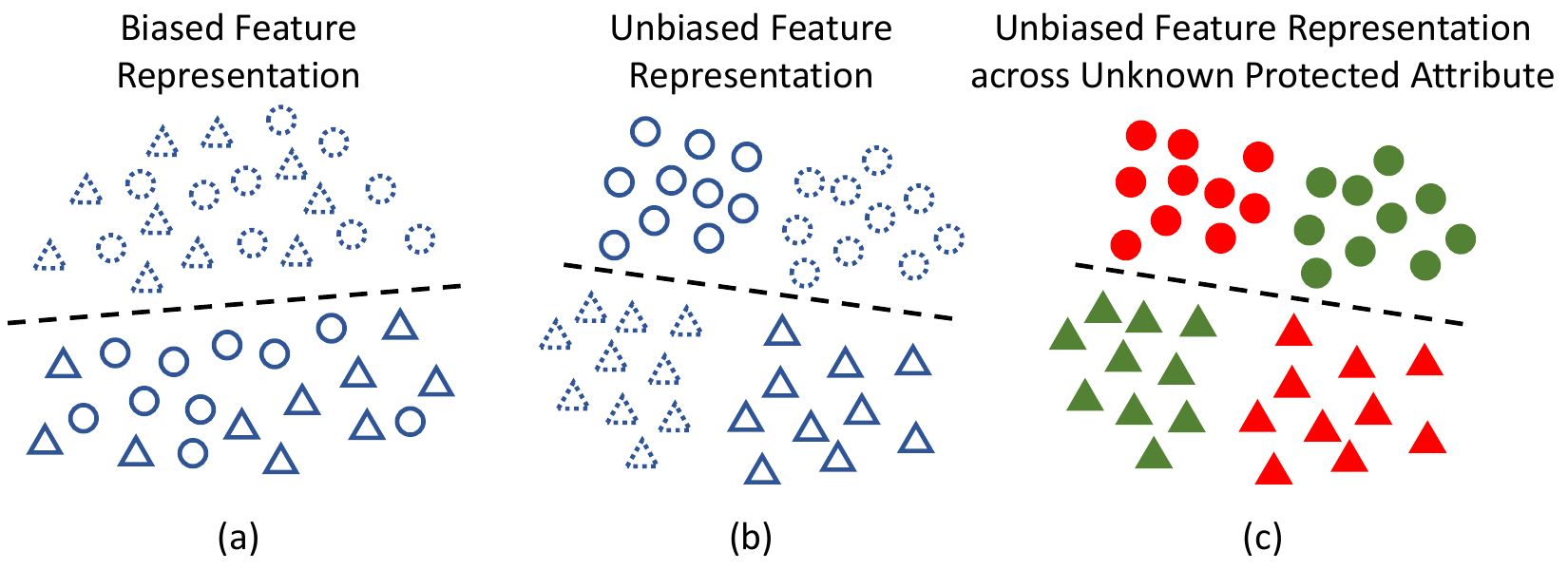}
\caption{A toy example of the feature visualization of a model trained for shape prediction on the testing set. \textit{Solid} and \textit{Dotted} represent classes of the non-protected attribute and \textit{Color} represents the protected attribute. (a) Shows the biased feature representations of the model (trained using conventional approaches). (b) Shows the unbiased feature representations of the model (trained using the proposed NPAD algorithm). (c) Shows the unbiased feature representations of the model across the unknown protected attribute (assuming the non-protected attribute is fully correlated with the protected attribute).}
\label{fig:FeatureVis}
\end{figure*}

\subsection{Bias Mitigation using Non-protected Attributes}
A convolutional neural network (CNN) model as a feature extractor is trained using the selected non-protected attributes for bias mitigation. The optimized model is further updated by adding dense layers for predicting attribute $\mathbf{A}_i$. The detailed process is discussed below.

Let $\Psi$ be the CNN model (feature extractor) to be optimized for bias mitigation and $\mathbf{N}$ be the set of selected non-protected attributes. The proposed NPAD algorithm combines the classes of each non-protected attribute in set $\mathbf{N}$ with the classes of $\mathbf{A}_i$ and selects the unique combinations. Each unique combination is considered as a separate class, resulting in $2^{n+1}$ classes (since attributes are binary) for model optimization. Here, $n$ is the number of non-protected attributes in set $\mathbf{N}$. Ideally, if the non-protected attribute is fully correlated with the protected attribute, then separating the clusters (formed by combining the classes of $\mathbf{A}_i$ along with the classes of the non-protected attribute) will increase the fairness of $\Psi$ across the protected attribute. Figure \ref{fig:FeatureVis} visually illustrates model fairness using non-protected attribute. Thus, the next step is to train $\Psi$ via optimizing the feature space with respect to the $2^{n+1}$ classes.

%In an ideal scenario if protected attribute information is available then one way to mitigate bias is to separate the clusters formed by combining the classes of attribute $\mathbf{A}_i$ along with the classes of protected attributes. Figure [] shows a visual illustration of the feature representation of an unbiased model. But, here we do not have the information of the protected attributes. Thus, we are optimizing the feature space using the clusters formed by combining the classes of attribute $\mathbf{A}_i$ and the non-protected attributes (that may be correlated with the protected attributes).

% unlearn the protected attribute information by separating

% It is observed that the feature space clearly separates the two classes of attribute $\mathbf{A}_i$ and is no longer separable by the protected attribute (demonstrating unbiased feature representations).

For optimizing the feature space, we propose the Debiasing via Attribute Cluster Loss (DACL) and Filter Redundancy Loss (FRL) functions. DACL provides supervision to the model to learn class-specific information for bias mitigation. It optimizes the clusters of each $2^{n+1}$ class by minimizing the intra-class distance and maximizing the inter-class distance. The intra-class distance is minimized by reducing the moving standard deviation of the feature representations of each class. Whereas, the inter-class distance is maximized by increasing the distance between the moving average of the feature representations of different classes. We use the moving average and standard deviation as it helps to reduce the effect of outliers. Moving average of the feature representation of class $p$ at iteration $t$ is:
\begin{equation}
\mathbf{M}_{p,t} = \frac{r_{p,t} \mathbf{Z}_{p,t} + r_{p,t-1} \mathbf{Z}_{p, t-1}}{r_{p,t} + r_{p, t-1}}
\end{equation}
Here, $\mathbf{Z}_{p,t}$ and $\mathbf{Z}_{p,t-1}$ are the mean feature representations of class $p$ at iteration $t$ and $t-1$, respectively. $r_{p,t}$ and $r_{p,t-1}$ are the number of samples in class $p$ at iteration $t$ and $t-1$, respectively. Moving standard deviation of the feature representation of class $p$ at iteration $t$ is:
\begin{multline}
    V_{p,t} = \Bigg[ \frac{r_{p,t}[\sigma_{p,t}^2 + (z_{p,t} - m_{p,t})^2]}{r_{p,t} + r_{p,t-1}} \quad + \\ \frac{r_{p,t-1}[\sigma_{p,t-1}^2 +  (z_{p,t-1} - m_{p,t})^2]}{r_{p,t} + r_{p,t-1}} \Bigg]^{1/2}
\end{multline}
where, $\sigma_{p,t}$ and $\sigma_{p,t-1}$ are the standard deviation of the feature representations of class $p$ at iteration $t$ and $t-1$, respectively. $z_{p,t}$ and $z_{p,t-1}$ are the mean of vectors $\mathbf{Z}_{p,t}$ and $\mathbf{Z}_{p,t-1}$, respectively. Similarly, $m_{p,t}$ and $m_{p,t-1}$ are the mean of vectors $\mathbf{M}_{p,t}$ and $\mathbf{M}_{p,t-1}$, respectively. DACL at iteration $t$ is mathematically represented as: 
\begin{equation}
    \mathcal{L}_{C,t} = \frac{\sum_{p=1}^{2^{n+1}} \sum_{q=1}^{2^{n+1}} V_{p,t} V_{q,t}}{||\mathbf{M}_{p,t} - \mathbf{M}_{q,t}||_2^2}  \quad \quad p \neq q
\end{equation}
For better separation of the $2^{n+1}$ clusters in feature space to mitigate bias in model prediction, it is important that the model $\Psi$ learn non-redundant features. However, in the conventional CNN learning process, it is observed that the channels learned for a specific filter are highly correlated with each other, which in turn increases the feature redundancy. To reduce redundancy and learn filters that enhance the feature expressiveness, we propose the Filter Redundancy Loss (FRL). FRL helps to boost the overall model performance by reducing the redundancy in the channels of a specific filter. Let $F^{H \times W \times C}$ be a filter and $\widehat{F}^{H \times W \times C}$ be the mean normalized filter. Mathematically, the FRL is:
\begin{equation}
    \mathcal{L}_F = \sum_{\forall a} \sum_{\forall b} |\widehat{F}^{H \times W \times C_a} \cdot \widehat{F}^{H \times W \times C_b}| \quad a \neq b
\end{equation}
%In the above loss function, the dot product among the channels of a specific filter is computed. This reduces the redundancy by increasing the de-correlation among the channels of the filter.

$\widehat F$ is obtained by subtracting $F_{mean}$ from $F$ to get the mean normalized filter. Here, we first compute the dot product between the mean normalized filters $\widehat F_a$ and $\widehat F_b$, then we take the mean across all the filters and minimize it. The FRL loss function reduces the redundancy by increasing the de-correlation among different learned filters. This will enhance the feature expressiveness and help to boost the overall model performance. The proposed FRL acts as a supplement to DACL for better separation of the clusters in the feature space, leading to fair model prediction. To reduce the computational complexity, we apply FRL on top-$k$ filters with the highest magnitude. The final loss, with $\lambda_1$ and $\lambda_2$ as hyperparameters, for bias mitigation is:
\begin{equation}
    \mathcal{L} = \lambda_1 \mathcal{L}_C + \lambda_2 \mathcal{L}_F
\end{equation}

\textbf{Bias-Invariant Attribute Prediction:} The DACL and FRL functions utilize the non-protected attributes to optimize the model for bias mitigation. The optimized model is used to predict the attribute $\mathbf{A}_i$ by adding dense layers after the final convolutional layer for bias-invariant prediction.

\section{Experimental Setup}
Experiments are performed for the task of facial attribute prediction and the performance is analyzed across two protected attributes (unknown during training). The details of the datasets with the corresponding protocols, details of the comparison algorithms, implementation details, and evaluation metrics are discussed below.

\subsection{Datasets and Protocols}
\noindent \textbf{LFWA dataset} \cite{huang2008labeled} contains $13,233$ images of $5,749$ subjects with $40$ annotated binary attributes. Standard pre-defined protocol \cite{huang2008labeled} is used for the experiments.

\noindent \textbf{CelebA dataset} \cite{liu2015faceattributes} contains $2,02,599$ face images of more than $10,000$ celebrities with $40$ annotated binary attributes. Standard pre-defined protocol \cite{liu2015faceattributes} is used.

%For both the datasets, \textit{male} and \textit{young} are considered as the protected attributes. $P_M$ and $P_{NM}$ denote \textit{male} and \textit{not male} (subgroups of \textit{gender}), respectively. $P_Y$ and $P_{NY}$ denote \textit{young} and \textit{not young} (subgroups of \textit{age}), respectively. The protected attributes are only used during evaluation.

\begin{figure}[]
\centering
\includegraphics[scale = 0.32]{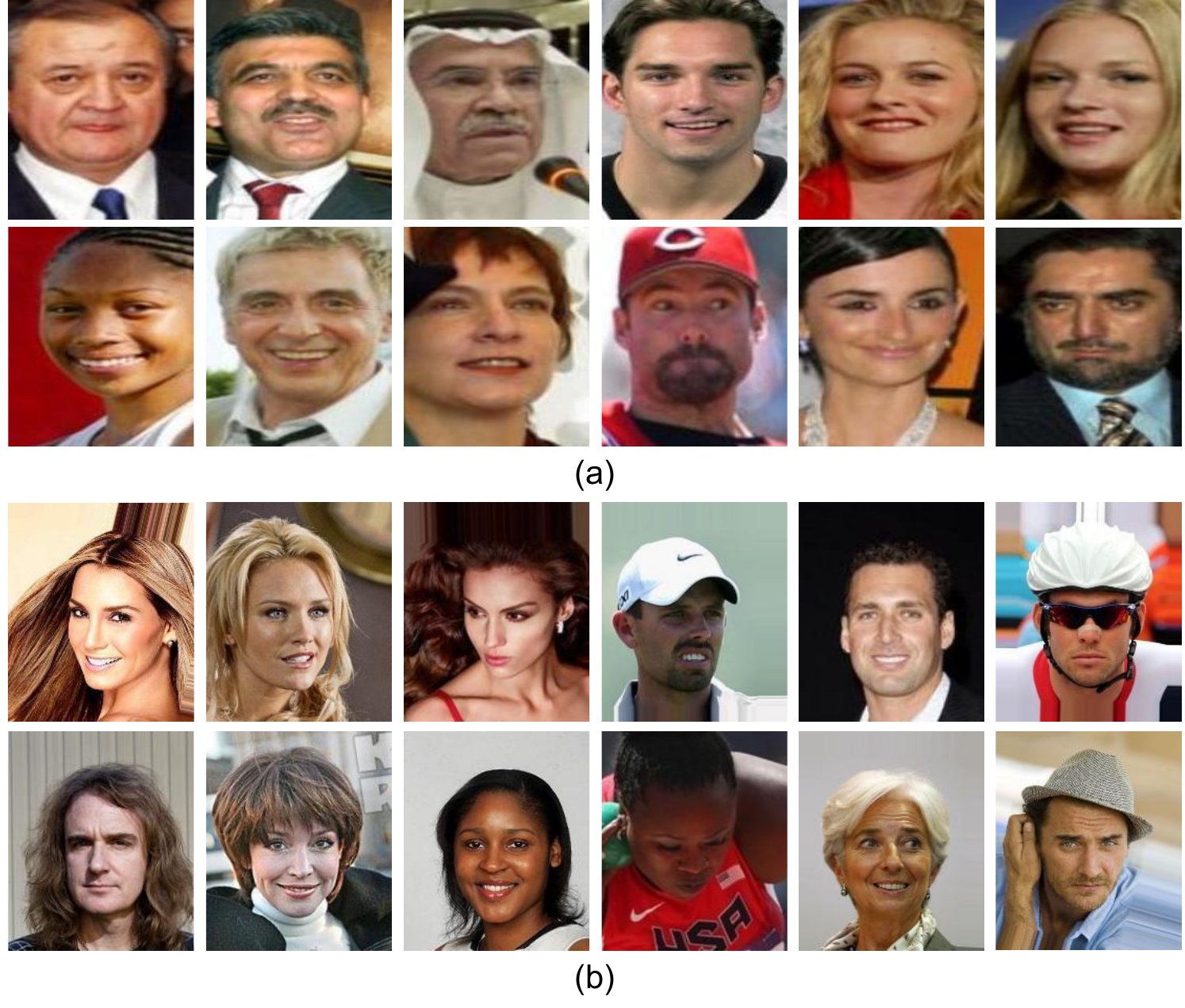}
\caption{Sample images of the (a) LFWA \cite{huang2008labeled} and (b) CelebA \cite{liu2015faceattributes} datasets. Images of both datasets are collected in unconstrained environmental settings with variation in pose, illumination, and the degree of occlusion.}
\label{fig:DatasetCollage}
\end{figure}

For both the datasets, \textit{male} and \textit{young} are considered as protected attributes. We denote $P_M$ and $P_{NM}$ as \textit{male} and \textit{not male}, respectively. $P_M$ and $P_{NM}$ are the subgroups of \textit{gender}. Similarly, $P_Y$ and $P_{NY}$ denote \textit{young} and \textit{not young}, respectively, which are considered as the subgroups of \textit{age}. The protected attributes are only used during evaluation. Sample images of the datasets are shown in Figure \ref{fig:DatasetCollage}.

\subsection{Details of the Comparison Algorithms}
For all the experiments, the model architecture consists of a CNN as a feature extractor followed by a two-layer neural network classifier (NNET). 

\textbf{Baseline Model Training (BMT):} In BMT, the last few convolutional layers and the NNET are trained for predicting attribute $\mathbf{A}_i$. The information related to the protected and other non-protected attributes is not used during BMT.

\textbf{Protected Attribute-based Debiasing (PAD) Algorithm:} In the PAD algorithm, we assume that the protected attribute corresponding to each input image is known during training. The CNN (feature extractor) is first trained using the proposed DACL and FRL functions. The optimized model is further trained by adding NNET for predicting attribute $\mathbf{A}_i$. This comparison is used to show the performance and fairness that can be achieved when the protected attributes are known. This sets a reference point for the proposed algorithm when the protected attributes are unknown.

\textbf{Learning from failure (LfF) \cite{nam2020learning}:} In LfF, a pair of neural networks is trained with two objectives: (i) amplifying bias in the first network and (ii) debiasing the second network by focusing on samples that go against
the prejudice of the first biased network. For a fair comparison, both networks follow similar architecture as BMT.

\textbf{Proposed NPAD Algorithm:} For model training using  NPAD algorithm, the selected non-protected attributes are utilized to optimize the CNN (feature extractor) using DACL and FRL. We perform the optimization under two scenarios. In the first scenario \textit{NPAD (One)}, one non-protected attribute is used for optimization whereas in the second scenario \textit{NPAD (Two)}, two non-protected attributes are used. In both the scenarios, the optimized model is further trained by adding NNET for predicting attribute $\mathbf{A}_i$.

%To the best of our knowledge, there is no work for bias mitigation without using protected attribute information in the vision domain. 

% LightCNN-29 model has shown high generalization abilities for face recognition. Thus, we have used this model and trained it for facial attribute prediction.

\subsection{Implementation Details}
LightCNN-29 \cite{wu2018light} architecture is used to perform the experiments. The model weights are initialized with those learned on the MS-Celeb-1M dataset \cite{guo2016ms}. To optimize the model using DACL and FRL, the model is trained for 10 epochs with a batch size of 200. Adam optimizer is used, and the learning rate is set to $0.0001$. $\lambda_1$ and $\lambda_2$ are set to 0.5. During training, the last 10 and 12 convolutional layers are updated for the LFWA and CelebA datasets, respectively. 

For attribute prediction, the optimized model is trained by adding NNET after the last convolutional layer using cross-entropy loss function. The dimensions of the layers in NNET are $128$ and $64$, respectively. Each layer is followed by ReLU activation. The optimized model $+$ NNET is trained for 10 epochs using SGD optimizer with $0.001$ learning rate. Momentum is set to 0.9 and a batch size of 50 is used. Code is implemented in PyTorch. All the experiments are performed on Nvidia GeForce GTX 1080 Ti.

\subsection{Evaluation Metrics}
Experimental results are reported using performance and bias evaluation metrics. Class-wise and overall classification accuracy are used for performance evaluation. Bias is measured using Degree of Bias (DoB) \cite{gong2019debface} and the proposed Overall Performance Equality (OPE) metric. DoB measures the standard deviation of classification accuracy across different subgroups of a protected attribute. The details of the OPE metric are discussed below.

\textbf{Overall Performance Equality (OPE):} Existing bias evaluation metrics such as Difference in Equality of Opportunity \cite{lokhande2020fairalm} and Predictive Parity \cite{garg2020fairness} measure the disparity in model performance across different subgroups. But, when the model predicts all the samples in one class, the disparity measures are unable to account for poor model performance. However, when the model predicts all the samples in one class, the disparity measures are unable to account for poor model performance. However, in the real world, an unbiased but low-performing model is undesirable. Therefore, we propose the OPE metric that considers the disparity in overall model performance across different subgroups.
\begin{equation}    
    OPE = |OP_{s_i} - OP_{s_j}|
\end{equation}
where, $OP_{s_i}$ and $OP_{s_j}$ represents the overall performance across subgroups $s_i$ and $s_j$, respectively. The overall performance for a particular subgroup $s_i$ is:
\begin{equation}
    OP_{s_i} = \\\frac{|tp - \hat{tp}| + \hat{fp} + \hat{fn} + |tn - \hat{tn}|}{2 \times Total Samples}
\end{equation}
where, $tp$ and $\hat{tp}$ represent the ground truth positives and predicted true positives, respectively. $\hat{fp}$ and $\hat{fn}$ represent the predicted false positives and false negatives, respectively. $tn$ and $\hat{tn}$ represent the ground truth negatives and predicted true negatives, respectively. Thus, $OP_{s_i}$ considers the number of false predictions along with the difference in the number of true predictions to estimate the overall performance of subgroup $s_i$. A lower value of DoB and OPE indicates low bias in model prediction.

\begin{table}[]
\centering
\small
\caption{Confusion matrix for `Bangs' prediction (Bangs as `B' and Not Bangs as `NB') across \textit{gender} subgroups on the LFWA dataset.}
\label{Tab:EvalMet}
\begin{tabular}{|c|c|c|c|c||c|c|}
\hline
 & \textbf{Algorithm}    &    & \multicolumn{2}{c||}{$P_M$ (\textbf{Predicted})}        & \multicolumn{2}{c|}{$P_{NM}$ (\textbf{Predicted})}           \\ \hline
% & \textbf{Algorithm}                   &    & \multicolumn{2}{c||}{\textbf{Predicted}} & \multicolumn{2}{c|}{\textbf{Predicted}} \\ \hline
\multirow{7}{*}{\rotatebox{90}{\textbf{Ground Truth}}} &                                      &    & B                  & NB                 & B                  & NB                 \\ \cline{2-7} 
 & \multirow{2}{*}{\textbf{BMT}}        & B  & 0                  & 766                & 0                  & 363                \\  
 &                                      & NB & 0                  & 4654               & 0                  & 1097               \\ \cline{2-7} 
 & \multirow{2}{*}{\textbf{NPAD (One)}} & B  & 553                & 213                & 292                & 71                 \\  
&                                      & NB & 253                & 4401               & 62                 & 1035               \\ \cline{2-7} 
& \multirow{2}{*}{\textbf{NPAD (Two)}} & B  & 511                & 255                & 289                & 74                 \\ 
 &                                      & NB & 202                & 4452               & 48                 & 1049               \\ \hline
\end{tabular}
\end{table}

Here, we provide an example to support our claim that existing bias evaluation metrics fail when all the samples are predicted in one class. Let us consider a bangs/not-bangs prediction model that predicts all the samples in one class (not bangs). In this scenario, the disparity in model prediction across different subgroups becomes zero, indicating an unbiased model. However, in the real world, an unbiased but low-performing model is undesirable. 

Table \ref{Tab:EvalMet} shows the confusion matrix for `Bangs' prediction across \textit{gender} subgroups using Baseline Model Training (BMT) approach and the proposed NPAD algorithm on the LFWA dataset. It is observed that BMT predicts all the samples as \textit{not bangs}. Therefore, the True Positive Rate (TPR) and False Positive Rate (FPR) becomes 0.00\%, while the True Negative Rate (TNR) and False Negative Rate (FNR) becomes 100.00\% for both \textit{genders}. The Difference in Equality of Opportunity (DEO) is calculated by taking the absolute difference between the FNR for both \textit{genders}. Predictive Parity is calculated by taking the absolute difference between the Positive Predictive Value (True Positives/(True Positives + False Positives)) for both \textit{genders}. Thus, in this case, DEO and PPV become 0.00\%, indicating an unbiased model. The DEO and PPV of the proposed NPAD (Two) algorithm is 12.90\% and 14.09\%, respectively. Here, the high DEO and PPV do not account for the improved model performance achieved by the proposed NPAD algorithm. This shows the limitations of the existing bias evaluation metrics. On the other hand, the proposed Overall Performance Equality (OPE) metric considers the disparity in overall model performance across different subgroups. In this example, OPE using BMT and NPAD (Two) are 10.73\% and 0.08\%, respectively, indicating a better evaluation metric that considers the overall model performance along with the disparity in model prediction across different subgroups. 

\begin{table*}[!t]
\centering
\renewcommand{\arraystretch}{1.4}
\caption{Performance of the proposed NPAD algorithm (\%) on three attributes of the LFWA dataset. The result closest to PAD is highlighted in bold.}
\label{Tab:LFWMain}
\resizebox{\columnwidth}{!}{
\begin{tabular}{cccccccccccc}
\hline
\multirow{3}{*}{\textbf{Attribute}}                                         & \multirow{3}{*}{\textbf{Algorithm}} & \multicolumn{5}{c}{\textbf{Across Gender}}                                                            & \multicolumn{5}{c}{\textbf{Across Age}}                                                               \\ \cline{3-12} 
&                                     & \multicolumn{3}{c}{\textbf{Accuracy $\uparrow$}} & \multirow{2}{*}{\textbf{DoB $\downarrow$}} & \multirow{2}{*}{\textbf{OPE $\downarrow$}} & \multicolumn{3}{c}{\textbf{Accuracy $\uparrow$}} & \multirow{2}{*}{\textbf{DoB $\downarrow$}} & \multirow{2}{*}{\textbf{OPE $\downarrow$}} \\ \cline{3-5} \cline{8-10}
         
&                                     & $P_M$  & $P_{NM}$  & \textbf{Overall}  &                               &                               & $P_Y$  & $P_{NY}$  & \textbf{Overall}  &                               &                               \\ \hline \hline
\multirow{4}{*}{Big Nose}                                                   & BMT                                 & 77.93  & 75.21     & 77.35             & 1.36                          & 2.73                          & 68.78  & 79.55     & 77.35             & 5.38                          & 10.77                         \\  
& PAD                                 & 83.21  & 82.47     & 83.05             & 0.37                          & 0.74                          & 80.26  & 84.15     & 83.36             & 1.94                          & 3.90                          \\  & LfF                                 & 81.01  & 77.94     & 80.36             & 1.53                          & 3.07                          & 75.19  & 81.69     & 80.36             & 3.24                          & 6.49                          \\
                                                                            & NPAD (One)                          & 82.99  & 81.92     & 82.76             & \textbf{0.53}                 & \textbf{1.07}                 & 78.62  & 83.82     & 82.76             & 2.60                          & 5.21                          \\  
                                                                            & NPAD (Two)                          & 82.95  & 80.95     & 82.52             & 1.00                          & 1.99                          & 78.75  & 83.49     & 82.52             & \textbf{2.37}                 & \textbf{4.73}                 \\ \hline 
\multirow{4}{*}{Black Hair}                                                 & BMT                                 & 88.76  & 83.42     & 87.63             & 2.67                          & 5.37                          & 77.83  & 90.14     & 87.63             & 6.15                          & 12.31                         \\  
                                                                            & PAD                                 & 91.92  & 89.18     & 91.34             & 1.37                          & 2.74                          & 84.18  & 92.79     & 91.03             & 4.30                          & 8.61                          \\  & LfF                                 & 91.35  & 87.05     & 90.44             & 2.15                          & 4.29                          & 82.89  & 92.37     & 90.44             & 4.73                          & 9.47                          \\ 
                                                                            & NPAD (One)                          & 91.72  & 88.29     & 90.99             & 1.71                          & 3.43                          & 83.54  & 92.90     & 90.99             & 4.68                          & 9.36                          \\  
                                                                            & NPAD (Two)                          & 91.69  & 88.97     & 91.11             & \textbf{1.36}                 & \textbf{2.72}                 & 84.39  & 92.84     & 91.11             & \textbf{4.22}                 & \textbf{8.45}   \\ \hline 
\multirow{4}{*}{\begin{tabular}[c]{@{}c@{}}5 o'clock\\ Shadow\end{tabular}} & BMT                                 & 66.83  & 93.90     & 72.57             & 13.53                         & 27.08                         & 82.82  & 69.95     & 72.57             & 6.43                          & 12.88                         \\  
& PAD                                 & 71.09  & 93.90     & 75.93             & 11.40                         & 22.82                         & 84.89  & 73.38     & 75.73             & 5.75                          & 11.51                         \\   & LfF                                 & 67.14  & 94.25     & 72.89             & 13.55                         & 27.11                        & 82.75  & 70.37     & 72.89             & 6.19                          & 12.38                         \\

                                                                            & NPAD (One)                          & 70.79  & 93.90     & 75.70             & 11.55                         & 23.11                         & 84.89  & 73.34     & 75.70             & \textbf{5.77}                 & \textbf{11.55}                \\  
                                                                            & NPAD (Two)                          & 70.63  & 93.49     & 75.48             & \textbf{11.43}                & \textbf{22.87}                & 85.25  & 72.98     & 75.48             & 6.13                          & 12.27                         \\ \hline
\end{tabular}}

\end{table*}

\begin{figure}[]
\centering
\includegraphics[scale = 0.5]{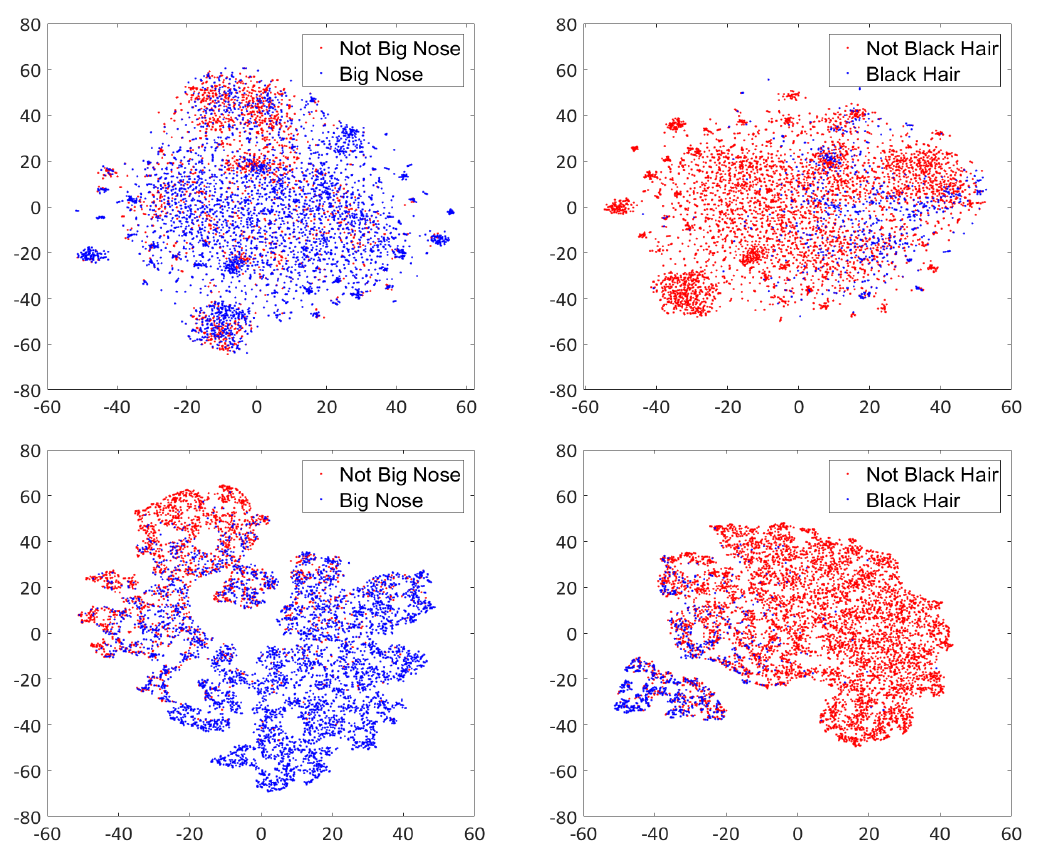}
\caption{t-SNE visualization using BMT (top row) and the proposed NPAD algorithm (bottom row) corresponding to attributes `Big Nose' and `Black Hair' of the LFWA dataset.}
\label{fig:TSNE}
\end{figure}

\section{Results and Analysis}
In order to evaluate the effectiveness of the proposed NPAD algorithm, experiments are performed on the LFWA and CelebA datasets. For the LFWA dataset, experimental results are shown corresponding to the `Big Nose', `Black Hair', and `5 o'çlock Shadow' attributes in Table \ref{Tab:LFWMain}. It is observed that PAD, LfF, and the proposed NPAD algorithms outperform BMT for all the attributes. For instance, the overall performance of PAD, LfF, NPAD (One), and NPAD (Two) increases by 6.01\%, 3.01\% 5.41\%, and 5.17\% compared to BMT corresponding to the `Big Nose' attribute across \textit{age} subgroups. Additionally, these algorithms show a significant reduction in DoB and OPE compared to BMT, indicating enhanced fairness in model prediction. It is important to note that the PAD algorithm is optimized using the proposed DACL and FRL functions, which leads to enhanced performance over BMT. 

While comparing NPAD with LfF, it is observed that NPAD outperforms LfF in terms of fairness and model performance. LfF is based on failure-based debiasing scheme, where debiased training is achieved by focusing on the samples that go against the prejudice of a biased network. On the other hand, NPAD algorithm utilizes the correlation of the protected attribute with the non-protected attribute for debiasing. The enhanced performance of NPAD over LfF shows the effectiveness of non-protected attributes in bias mitigation. Figure \ref{fig:TSNE} shows the t-Distributed Stochastic Neighbor Embedding (t-SNE) visualization of the feature representations of the last convolutional layers using BMT and the proposed NPAD algorithm corresponding to the `Big Nose' and `Black Hair' attributes of the LFWA dataset. It is observed that the overlap among the classes is reduced using the proposed NPAD algorithm. Table \ref{Tab:CelebAMain} shows the results on the CelebA dataset corresponding to the `Wearing Necktie', `Sideburns', and `Gray Hair' attributes. A similar set of observations is drawn regarding the model fairness and the overall performance.

\begin{table*}[]
\centering
\renewcommand{\arraystretch}{1.4}
\caption{Performance of the proposed NPAD algorithm (\%) on three attributes of the CelebA dataset. The result closest to PAD is highlighted in bold.}
\label{Tab:CelebAMain}
\resizebox{\columnwidth}{!}{
\begin{tabular}{cccccccccccc}
\hline
\multirow{3}{*}{\textbf{Attribute}}                                        & \multirow{3}{*}{\textbf{Algorithm}} & \multicolumn{5}{c}{\textbf{Across Gender}}                                                            & \multicolumn{5}{c}{\textbf{Across Age}}                                                               \\ \cline{3-12} 
                                                                           &                                     & \multicolumn{3}{c}{\textbf{Accuracy $\uparrow$}} & \multirow{2}{*}{\textbf{DoB $\downarrow$}} & \multirow{2}{*}{\textbf{OPE $\downarrow$}} & \multicolumn{3}{c}{\textbf{Accuracy $\uparrow$}} & \multirow{2}{*}{\textbf{DoB $\downarrow$}} & \multirow{2}{*}{\textbf{OPE $\downarrow$}} \\ \cline{3-5} \cline{8-10}
                                                                           &                                     & $P_M$  & $P_{NM}$  & \textbf{Overall}  &                               &                               & $P_Y$  & $P_{NY}$  & \textbf{Overall}  &                               &                               \\ \hline \hline
\multirow{4}{*}{\begin{tabular}[c]{@{}c@{}}Wearing\\ Necktie\end{tabular}} & BMT                                 & 81.97  & 99.93     & 92.99             & 8.98                          & 17.96                         & 96.86  & 80.94     & 92.99             & 7.96                          & 15.92                         \\  
                                                                           & PAD                                 & 91.29  & 99.90     & 96.57             & 4.30                          & 8.61                          & 98.25  & 92.53     & 96.86             & 2.86                          & 5.71                          \\  & LfF                                 & 81.97  & 99.93     & 92.99             & 8.98                          & 17.96                         & 96.86  & 80.94     & 92.99             & 7.96                          & 15.92                         \\ 
                                                                           & NPAD (One)                          & 90.71  & 99.89     & 96.34             & 4.59                          & 9.19                          & 97.93  & 91.40     & 96.34             & 3.26                          & 6.53                          \\  
                                                                           & NPAD (Two)                          & 91.51  & 99.91     & 96.66             & \textbf{4.20}                 & \textbf{8.40}                 & 98.23  & 91.79     & 96.66             & \textbf{3.22}                 & \textbf{6.44}                 \\ \hline 
\multirow{4}{*}{Sideburns}                                                 & BMT                                 & 88.01  & 99.99     & 95.36             & 5.99                          & 11.98                         & 96.23  & 92.66     & 95.36             & 1.78                          & 3.57                          \\  
                                                                           & PAD                                 & 94.00  & 99.99     & 97.68             & 2.99                          & 5.99                          & 97.90  & 96.27     & 97.50             & 0.81                          & 1.63                          \\ & LfF                                 & 88.01  & 99.99     & 95.36             & 5.99                          & 11.98                         & 96.23  & 92.66     & 95.36             & 1.78                          & 3.57                          \\  
                                                                           & NPAD (One)                          & 93.83  & 99.99     & 97.61             & \textbf{3.08}                 & \textbf{6.16}                 & 98.03  & 96.31     & 97.61             & \textbf{0.86}                 & \textbf{1.72}                 \\  
                                                                           & NPAD (Two)                          & 93.16  & 99.99     & 97.35             & 3.41                          & 6.84                          & 97.82  & 95.9      & 97.35             & 0.96                          & 1.92                          \\ \hline 
\multirow{4}{*}{Gray Hair}                                                 & BMT                                 & 93.18  & 99.10     & 96.81             & 2.96                          & 5.92                          & 99.84  & 87.38     & 96.81             & 6.23                          & 12.46                         \\  
                                                                           & PAD                                 & 96.02  & 99.28     & 98.02             & 1.63                          & 3.26                          & 99.79  & 92.51     & 98.02             & 3.64                          & 7.28                          \\  & LfF                                 & 93.18  & 99.10     & 96.81             & 2.96                          & 5.92                          & 99.84  & 87.38     & 96.81             & 6.23                          & 12.46                         \\
                                                                           & NPAD (One)                          & 95.91  & 99.26     & 97.97             & \textbf{1.67}                 & \textbf{3.35}                 & 99.81  & 92.22     & 97.97             & \textbf{3.79}                 & \textbf{7.59}                 \\  
                                                                           & NPAD (Two)                          & 95.63  & 99.23     & 97.84             & 1.80                          & 3.60                          & 99.80  & 91.72     & 97.84             & 4.04                          & 8.07                          \\ \hline
\end{tabular}}
\end{table*}

On comparing the performance of PAD and NPAD algorithms, it is found that NPAD gives comparable performance and, in some cases, outperforms the PAD algorithm. It is further observed that NPAD (Two) is able to achieve better performance than PAD. It is our assertion that in such cases, the combination of two non-protected attributes provides better supervision to the model for bias mitigation. Tables \ref{Tab:LFWMain} and \ref{Tab:CelebAMain} also indicates that in some cases NPAD (One) is performing better than NPAD (Two). We assert that these are the cases where the first non-protected attribute is correlated with the protected attribute, and since the second non-protected attribute is independent of the first attribute, it indirectly penalizes the fairness and model performance. To validate this, we compute the correlation between non-protected attributes used for debiasing the models with the protected attribute `Gender.' Since all the attributes are categorical, we used Cramér's V method to find the correlation. From Table \ref{Tab:Corr}, it is observed that for `Wearing Necktie' model, the second non-protected attribute is more correlated with `Gender' and thus helps to boost the fairness and model performance using NPAD (Two). On the other hand, for `Sideburns' and `Gray Hair' models, the second non-protected attribute is less correlated with the protected attribute, which indirectly penalizes the fairness and model performance.

\begin{table}[]
\centering
\renewcommand{\arraystretch}{1.2}
\caption{Correlation between the non-protected attributes used for debiasing with the protected attribute `Gender.'}
\label{Tab:Corr}
\resizebox{\columnwidth}{!}{
\begin{tabular}{ccc}
\hline
\textbf{Attribute}       & \textbf{\begin{tabular}[c]{@{}c@{}}First Non-protected Attribute\end{tabular}} & \textbf{\begin{tabular}[c]{@{}c@{}}Second Non-protected Attribute\end{tabular}} \\ \hline
\textbf{Wearing Necktie} & 0.19                                                                              & 0.42                                                                               \\
\textbf{Sideburns}       & 0.79                                                                              & 0.02                                                                               \\
\textbf{Gray Hair}       & 0.11                                                                              & 0.07                                                                               \\ \hline
\end{tabular}}
\end{table}

\begin{figure*}[]
\centering
\includegraphics[scale = 0.45]{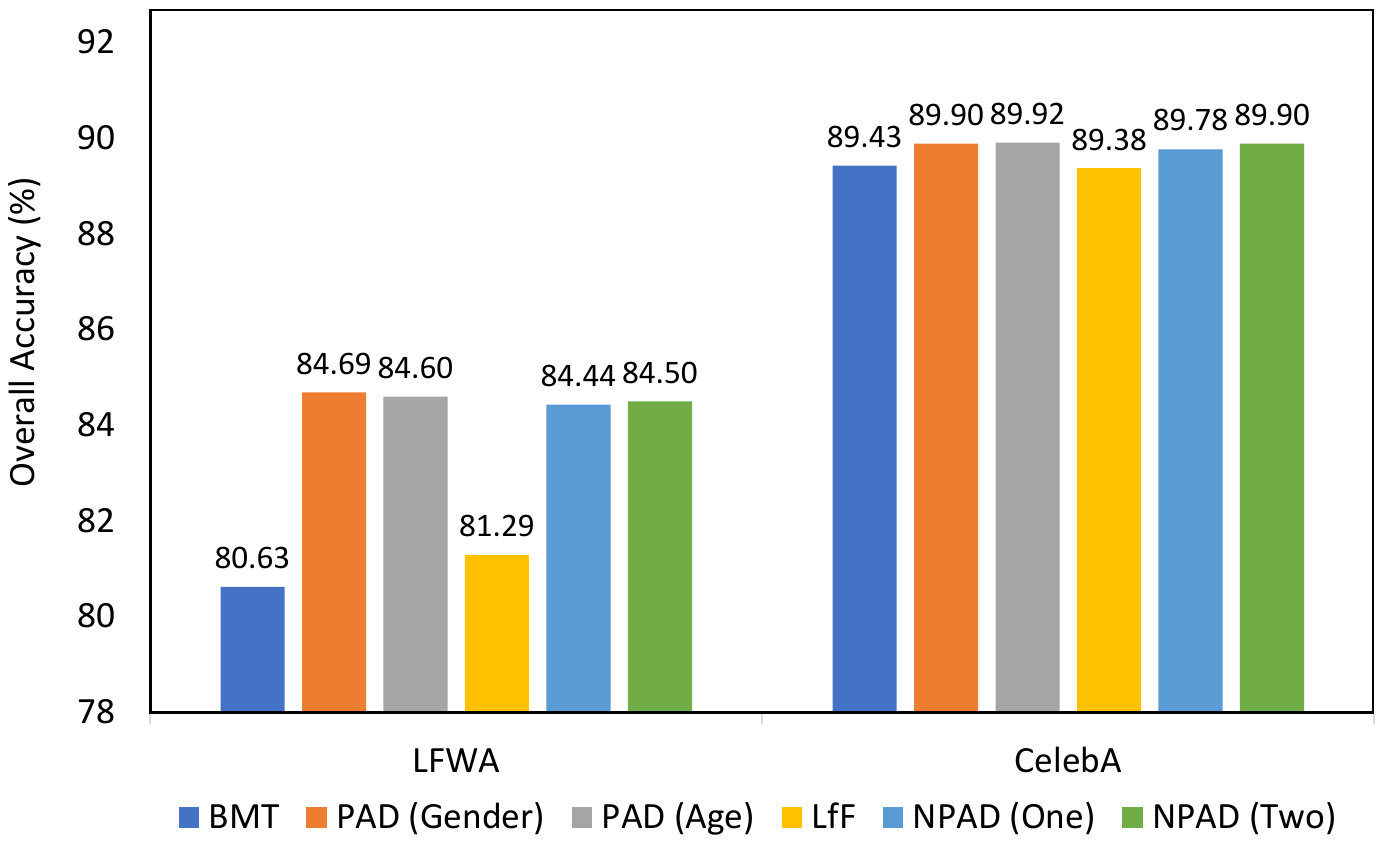}
\caption{Mean overall classification accuracy corresponding to all the attributes (excluding \textit{male} and \textit{young}) using BMT, PAD, LfF, and NPAD algorithms on the LFWA and CelebA datasets. PAD (Gender) and PAD (Age) indicates the performance when optimization is done using \textit{gender} and \textit{age} information, respectively.}
\label{fig:All_Att_Accuracy}
\end{figure*}

\begin{figure*}[!t]
\centering
\includegraphics[width=\textwidth,height=\textheight,keepaspectratio]{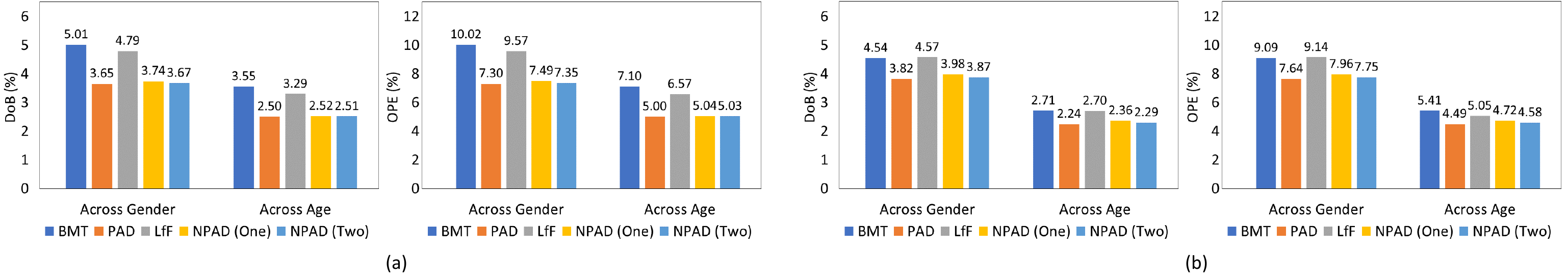}
\caption{Mean DoB and mean OPE corresponding to all the attributes (excluding \textit{male} and \textit{young}) using BMT, PAD, LfF, and NPAD algorithm on the (a) LFWA and (b) CelebA datasets.}
\label{fig:All_Att_DoBOPE}
\end{figure*}

The mean overall classification accuracy, DoB, and OPE corresponding to all the $40$ attributes of both the datasets are shown in Figures \ref{fig:All_Att_Accuracy} and \ref{fig:All_Att_DoBOPE}, respectively. The proposed NPAD algorithm achieves better performance and fairness compared to BMT. The accuracy, DoB, and OPE of the NPAD algorithm are close to the PAD algorithm. This showcases the effectiveness of the proposed NPAD algorithm to mitigate bias in model prediction in the absence of protected attribute information which further highlights the applicability of the NPAD algorithm in real-world applications.

The results and analysis shown above summarize the model fairness and performance across the subgroups of a single protected attribute (for instance \textit{Gender} or \textit{Age}). However, it is also important that the model must enhance performance and perform equally across the intersectional subgroups of multiple protected attributes (for instance, \textit\{Male, Young\}, \textit\{Female, Young\}, \textit\{Male, Old\}, \textit\{Female, Old\}). Therefore, we have also evaluated the performance of the proposed algorithm across the intersectional subgroups of \textit{gender} and \textit{age}. Results are compared with BMT and are shown in Table \ref{Tab:LFWInter}. It is observed that the proposed algorithm improves the performance and fairness of the model predictions. This shows that the model trained using the proposed NPAD algorithm performs well across the intersectional subgroups.

\begin{table}[]
\centering
\caption{Mean OPE (\%) for all the attributes (excluding male and young) on the LFWA and CelebA datasets across the intersectional subgroups of \textit{gender} and \textit{age}.}
\label{Tab:LFWInter}
\begin{tabular}{cccc}
\hline
       & BMT  & NPAD (One) & NPAD (Two) \\ \hline
LFWA   & 8.64 & 6.40       & 6.25       \\
CelebA & 8.39 & 7.12       & 7.03       \\ \hline
\end{tabular}
\end{table}

\begin{figure}[!t]
\centering
\includegraphics[scale = 0.35]{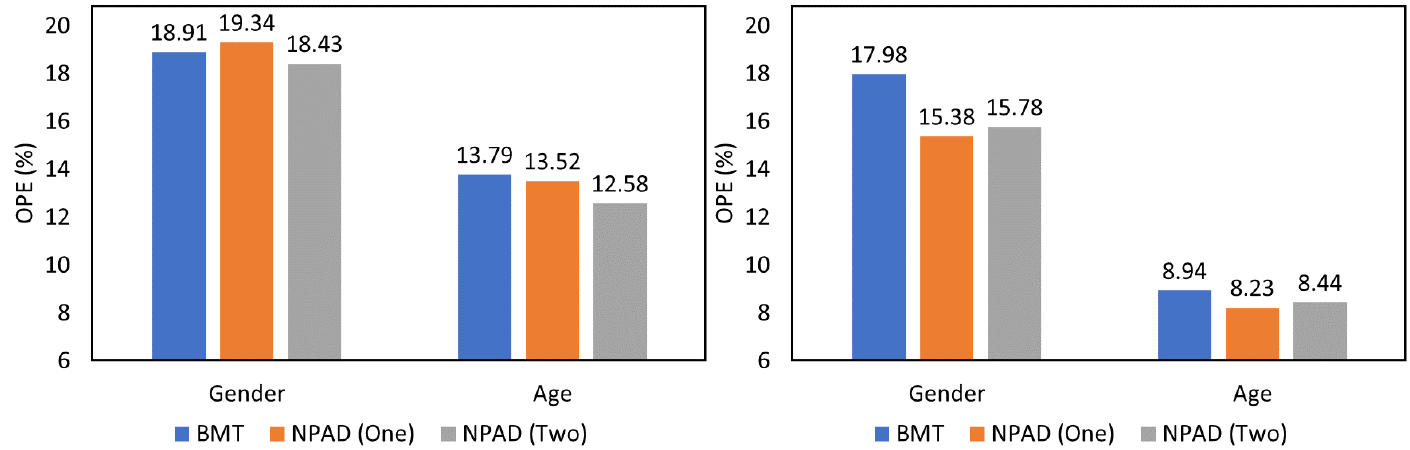}
\caption{Cross dataset experiment: mean OPE (\%) corresponding to all the attributes (excluding male and young) using BMT and NPAD on (a) LFWA and (b) CelebA datasets.}
\label{fig:CrossDb}
\end{figure}

To analyze the generalizability of the NPAD algorithm, experiments are performed in cross-dataset settings, where the models trained on the LFWA dataset are evaluated on the CelebA dataset and vice-versa. The results of the cross-dataset experiment are shown in Figure \ref{fig:CrossDb}. It is observed that the proposed algorithm achieves the lowest OPE. This shows the generalizability of the proposed algorithm across different datasets. Also, we observe that NPAD (One) performs better on the CelebA dataset compared to NPAD (Two). Model training using NPAD (Two) requires more classes for optimization compared to NPAD (One) which leads to less number of samples per class for NPAD (Two) during training. Since the LFWA dataset is a small dataset, we assert that reducing the number of samples per class negatively affects the model performance in cross-dataset experimental settings. Therefore, in this case NPAD (One) outperforms NPAD (Two).

\begin{figure*}[]
\centering
\includegraphics[scale = 0.64]{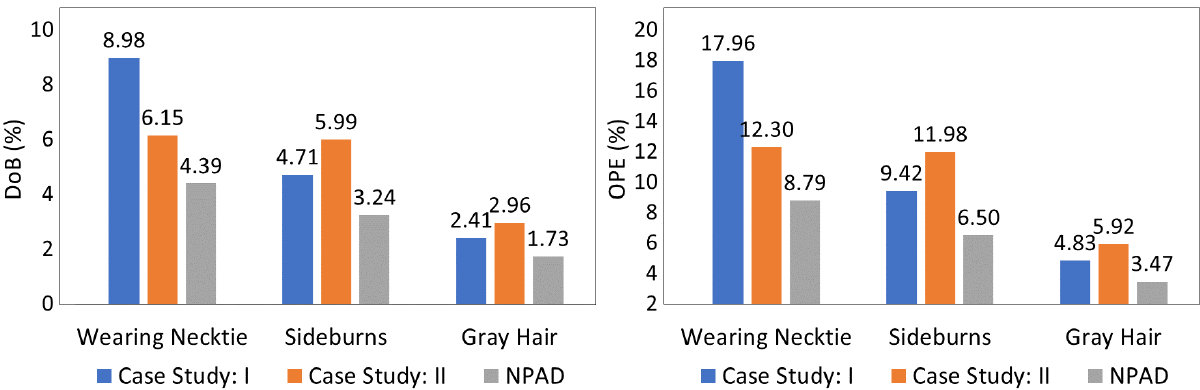}
\caption{Mean DoB and mean OPE of the case studies corresponding to the three attributes of the CelebA dataset across \textit{gender} subgroups. Comparison is performed with mean NPAD.}
\label{fig:CaseStudyCeleba_DoBOPE}
\end{figure*}

\begin{figure*}[t]
\centering
\includegraphics[scale = 0.4]{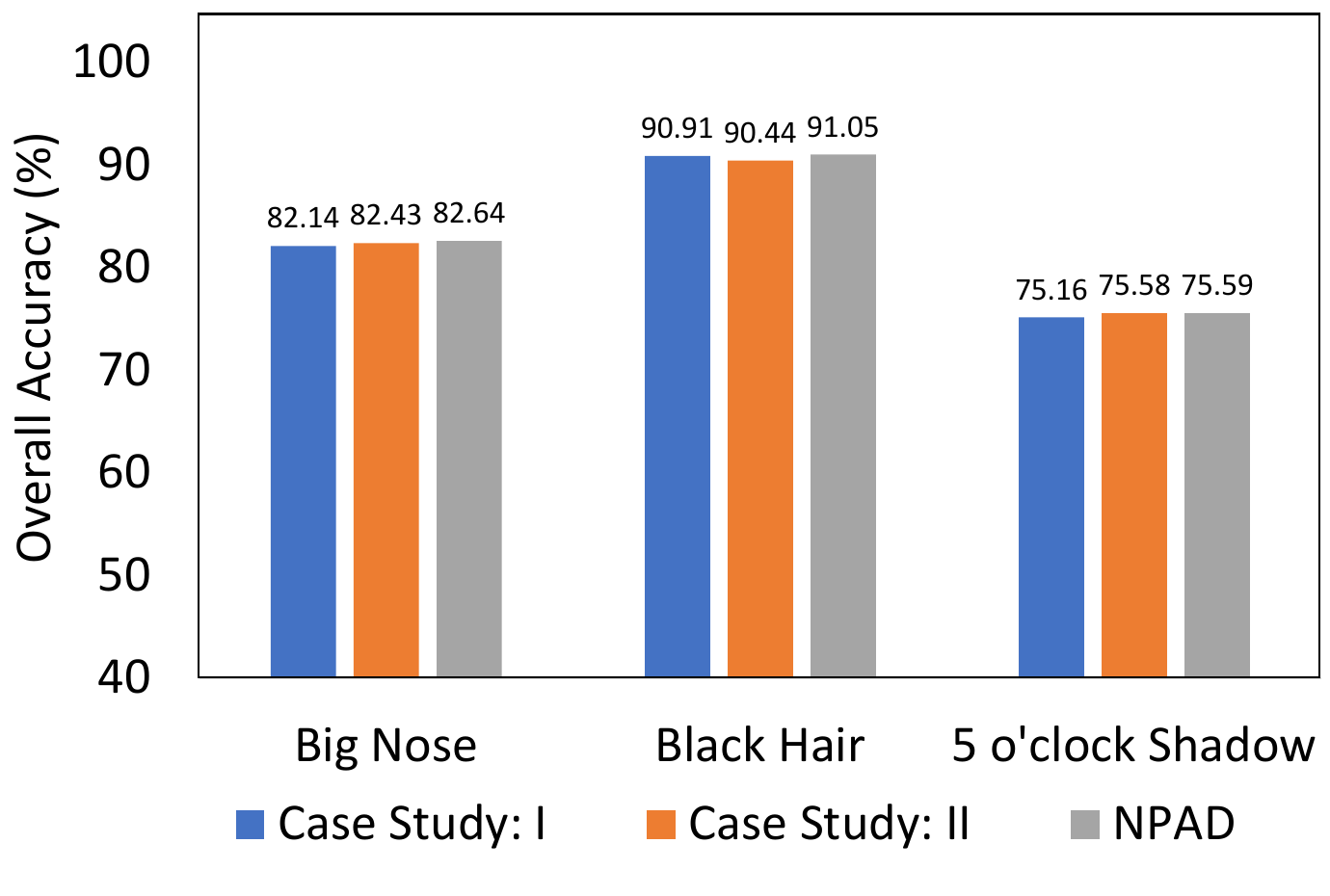}
\caption{Overall accuracy of the case studies corresponding to the three attributes of the LFWA dataset. Comparison is performed with mean NPAD.}
\label{fig:CaseStudyLFWAAcc}
\end{figure*}

\noindent \textbf{Case Study I: Random Attribute Selection} In this case study, the non-protected attributes are randomly selected from set $\mathbf{A}$ to optimize the model using DACL and FRL functions. The optimized model is further trained by adding NNET for predicting attribute $\mathbf{A}_i$. Two different experiments are performed (i) Case Study: I (One) and (ii) Case Study: I (Two). In the first experiment, one attribute is randomly selected from set $\mathbf{A}$ (excluding $\mathbf{A}_i$), while in the second experiment, two attributes are randomly selected from set $\mathbf{A}$ (excluding $\mathbf{A}_i$) for model optimization. The aim of the experiment is to analyze the model performance and fairness when the non-protected attributes are not selected intelligently. Figure \ref{fig:CaseStudyCeleba_DoBOPE} shows the mean DoB and OPE of the two experiments corresponding to the three attributes of the CelebA dataset across \textit{gender} subgroups. Comparison is done with the proposed NPAD algorithm. As observed from Figure \ref{fig:CaseStudyCeleba_DoBOPE}, NPAD achieves the lowest DoB and OPE. Figure \ref{fig:CaseStudyLFWAAcc} shows the mean overall accuracy of the two experiments corresponding to the three attributes of the LFWA dataset. Here also NPAD achieves the best overall accuracy. This showcases that intelligently selecting the non-protected attributes significantly reduces bias in model prediction and enhances the overall model performance.

\noindent \textbf{Case Study II: Without Protected/Non-Protected Attribute Information} This case study assumes that the information related to the protected and other non-protected attributes corresponding to each input image is unavailable. The model is optimized using DACL and FRL functions to separate the clusters formed by the classes of attribute $\mathbf{A}_i$. The optimized model is further trained by adding NNET for predicting attribute $\mathbf{A}_i$. This experiment aims to analyze the role of non-protected attributes in bias mitigation and boosting the overall model performance. DoB and OPE corresponding to the three attributes of the CelebA dataset across \textit{gender} subgroups are shown in Figure \ref{fig:CaseStudyCeleba_DoBOPE}. The high DoB and OPE indicate that the model is unable to mitigate bias in model prediction in the absence of non-protected attribute information. Figure \ref{fig:CaseStudyLFWAAcc} shows the overall model performance corresponding to the three attributes of the LFWA dataset. It is observed that the proposed NPAD algorithm achieves the best overall accuracy. The auxiliary information provided by the non-protected attributes help to boost the model performance further.

\begin{table}[]
\centering
\renewcommand{\arraystretch}{1.2}
\caption{Performance of the model (\%) by ablating individual loss terms of the NPAD algorithm (one attribute) for `Big Nose' prediction on the LFWA dataset across different \textit{gender} subgroups.}
\label{Tab:AblationStudy}
\begin{tabular}{cccccc}
\hline
\multirow{2}{*}{\textbf{\begin{tabular}[c]{@{}c@{}}Loss\end{tabular}}} & \multicolumn{3}{c}{\textbf{Accuracy $\uparrow$}} & \multirow{2}{*}{\textbf{DoB $\downarrow$}} & \multirow{2}{*}{\textbf{OPE $\downarrow$}} \\ \cline{2-4}
                                                                              & $P_M$   & $P_{NM}$   & Overall         &                               &                               \\ \hline \hline
DACL                                                                           & 82.86   & 81.23      & 82.51           & 0.81                          & 1.63                          \\ 
FRL                                                                            & 78.93   & 76.10      & 78.33           & 1.41                          & 2.83                          \\ 
Proposed                                                             & 82.99   & 81.92      & \textbf{82.76}  & \textbf{0.53}                 & \textbf{1.07}                 \\ \hline
\end{tabular}
\end{table}

\begin{figure}[]
\centering
\includegraphics[scale = 0.40]{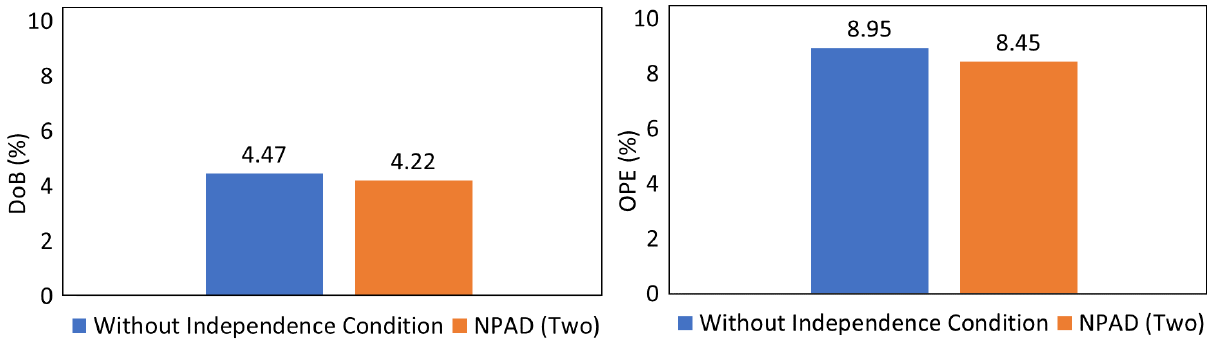}
\caption{DoB and OPE of the model when the independence between non-protected attributes is not considered. The performance is shown for `Black Hair' prediction across \textit{age} subgroups on the LFWA dataset.}
\label{fig:Ind_Dob_OPE}
\end{figure}

\noindent \textbf{Ablation Study:} Experiments are performed by ablating different loss terms of the NPAD algorithm one at a time during training to analyze the effectiveness of individual loss terms. Results are summarized in Table \ref{Tab:AblationStudy}. It is observed that on ablating DACL, the overall accuracy degrades significantly to 78.33\% and the bias in model prediction increases sharply. The drop in overall performance is because FRL does not provide any class-specific information. It only reduces the redundancy in the filter channels to learn non-redundant features. On the other hand, using only DACL shows better overall accuracy and fairness as it provides class-specific information to separate the clusters for bias mitigation. However, when DACL is combined with FRL, the overall performance and model fairness are further enhanced. This highlights the importance of FRL in the proposed NPAD algorithm. Additionally, an experiment is performed to analyze the effect of bias if the independence condition is not used during non-protected attribute selection. In other words, the selected non-protected attributes are dependent on each other. Figure \ref{fig:Ind_Dob_OPE} shows the results for ‘Black Hair’ prediction across \textit{age} subgroups on the LFWA dataset. It is observed that fairness in model prediction reduces when dependent non-protected attributes are used for optimization.

\begin{table}[]
\centering
\renewcommand{\arraystretch}{1.2}
\caption{Comparison with existing algorithm on the LFWA dataset. Average over all attributes (excluding protected attributes - male and young) is reported.}
\label{Tab:SOTA}
\begin{tabular}{cccccc}
\hline
\multirow{2}{*}{Algo} & \multicolumn{2}{c}{DoB $\downarrow$}       & \multicolumn{2}{c}{OPE $\downarrow$}       & \multirow{2}{*}{\begin{tabular}[c]{@{}c@{}}Overall\\ Accuracy $\uparrow$ \end{tabular}} \\ \cline{2-5}
                           & Gender        & Age           & Gender        & Age           &                                                                             \\ \hline \hline
ARL                       & 4.70          & 3.34          & 9.41          & 6.68          & 80.33                                                                       \\
Proposed                & \textbf{3.67} & \textbf{2.51} & \textbf{7.35} & \textbf{5.03} & \textbf{84.50}                                                              \\ \hline
\end{tabular}
\end{table}

\subsection{Additional Comparison with existing algorithm}
We have compared our proposed algorithm with Adversarially Reweighted Learning (ARL)~\cite{lahoti2020fairness}. The ARL algorithm does not use protected attribute information for bias mitigation. ARL focuses on improving worst-case protected groups (unknown) by leveraging the notion of computationally identifiable errors. They assume that optimizing the model for the worst-performing group will improve the model's fairness. In contrast, our proposed NPAD algorithm focuses on reducing the disparity between all the protected groups by optimizing the feature space utilizing the non-protected attribute information. We positively exploit the correlation between the non-protected and protected attributes for bias mitigation.

For the experiments, feature embeddings obtained from baseline models corresponding to the non-protected attributes are provided as input to the ANN~\cite{lahoti2020fairness}, which is then trained using the ARL algorithm. Table \ref{Tab:SOTA} summarizes the result corresponding to all the attributes on the LFWA dataset. It is observed that NPAD algorithm outperforms ARL. ARL focuses on improving worst-case protected groups (unknown) by leveraging the notion of computationally-identifiable errors. In contrast, our algorithm focuses on reducing the disparity between all the protected groups by optimizing the feature space utilizing the non-protected attribute information. We assert that optimizing for all the groups leads to better performance and fairer models.

\section{Conclusion and Discussion}
%How many non-protected attributes we need to select is an open challenge.
%Weightage of the non-protected attributes is not known.
% The presence of bias is an important problem in deep learning, and several algorithms have been proposed for its mitigation using protected attribute information. 
This research addresses the problem of bias mitigation in the absence of protected attribute information. The protected attributes are not always available and the legal and regulatory norms prohibit the use of protected attribute information in making important decisions. In such scenarios, non-protected attribute information can be utilized for model debiasing. Most of the publicly available datasets are labeled with non-protected attributes but do not have the protected attribute information. Hand-labeling large datasets for attribute information is time-consuming, unscalable, and requires manual effort. Further, annotating images for protected attributes (e.g. race) is more challenging compared to non-protected attributes (e.g. wearing\_glasses). Thus, utilizing non-protected attribute information for bias mitigation is beneficial.

%These datasets can be used to learn fair models that produce unbiased outcomes across different protected groups.
To achieve model fairness in the absence of protected attribute information, we intelligently utilize the information provided by a subset of the non-protected attributes. The proposed \textit{NPAD} algorithm does not use gender (or age) information for bias mitigation, or attempt to perform gender (or age) prediction. NPAD optimizes the feature space by separating the clusters formed using $2^{n+1}$ classes with the proposed DACL and FRL functions. DACL optimizes the model for fairness by ensuring effective learning in the feature space, while FRL maximizes learning of non-redundant features. The optimization with $2^{n+1}$ classes using the NPAD algorithm may seem computationally expensive but several problems in computer vision involve optimizing large numbers of classes. For example, in face recognition, deep models optimize feature space to separate embeddings of thousands of identities (classes). Similarly, the proposed NPAD learns discriminative features to separate the cluster formed using $2^{n+1}$ classes. The experiments and case studies show that the NPAD algorithm effectively reduces bias across protected attributes and enhances the overall model performance. This highlights that the information provided by the non-protected attributes, when used efficiently, can reduce bias and improve the model's fairness. However, the selection of an optimal number of non-protected attributes for mitigation remains an open problem. In the current experimental setup, the proposed NPAD algorithm is benchmarked on binary attributes. As a part of future work, the proposed NPAD algorithm can be extended to non-binary attributes.

% \bibliographystyle{elsarticle-num} 
% \bibliography{mybibfile}
\bibliography{references}
\end{document}